\documentclass[letterpaper, 10 pt, conference]{ieeeconf}  

\IEEEoverridecommandlockouts                              

\usepackage{graphics} 
\usepackage{epsfig} 
\usepackage{mathptmx} 
\usepackage{times} 
\usepackage{amsmath} 

\usepackage{amssymb}  
\usepackage{bm}
\usepackage{algorithm}
\usepackage[noend]{algpseudocode}
\usepackage{physics}
\usepackage{xcolor}
\usepackage{tikz}
\usetikzlibrary{fit,
                positioning}
\usetikzlibrary{calc} 
\usetikzlibrary{positioning}
\usepackage{balance}
\usepackage{soul}
\usetikzlibrary{arrows}
\usepackage{cite}
\usepackage{url}
\usepackage{multirow}
\usepackage{booktabs} 
\usepackage{listings}
\usepackage{hyperref}
\usepackage{afterpage}
\usepackage{algorithm}
\usepackage{algpseudocode}
\usepackage{setspace}
\usepackage{caption}
\usepackage{subcaption}
\definecolor{commentgreen}{rgb}{0.0, 0.5, 0.0} 
\title{\LARGE \bf
Multi-Robot Assembly of Deformable Linear Objects \\ Using Multi-Modal Perception
}

\author{Kejia Chen$^*$$^{1}$, Celina Dettmering$^*$$^{2}$, Florian Pachler$^{2}$, Zhuo Liu$^{1}$, Yue Zhang$^{1}$, Tailai Cheng$^{1}$,\\  Jonas Dirr$^{2}$, Zhenshan Bing$^{1}$, Alois Knoll$^{1}$, Rüdiger Daub$^{2,3}$
\thanks{\textsuperscript{*} Equal contribution. }
\thanks{$^{1}$ Chair of Robotics, Artificial Intelligence and Real-time Systems, School of Computation, Information and Technology, Technical University of Munich, Germany.}
\thanks{$^{2}$Institute for Machine Tools and Industrial Management, School of Engineering and Design, Technical University of Munich, Germany}
\thanks{$^{3}$Fraunhofer Institute for Casting, Composite and Processing Technology IGCV, Germany}
}

\let\oldtwocolumn\twocolumn
\renewcommand\twocolumn[1][]{%
    \oldtwocolumn[{#1}{
    \vspace{-20pt}
    \begin{center}
           \includegraphics[width=0.9\textwidth]{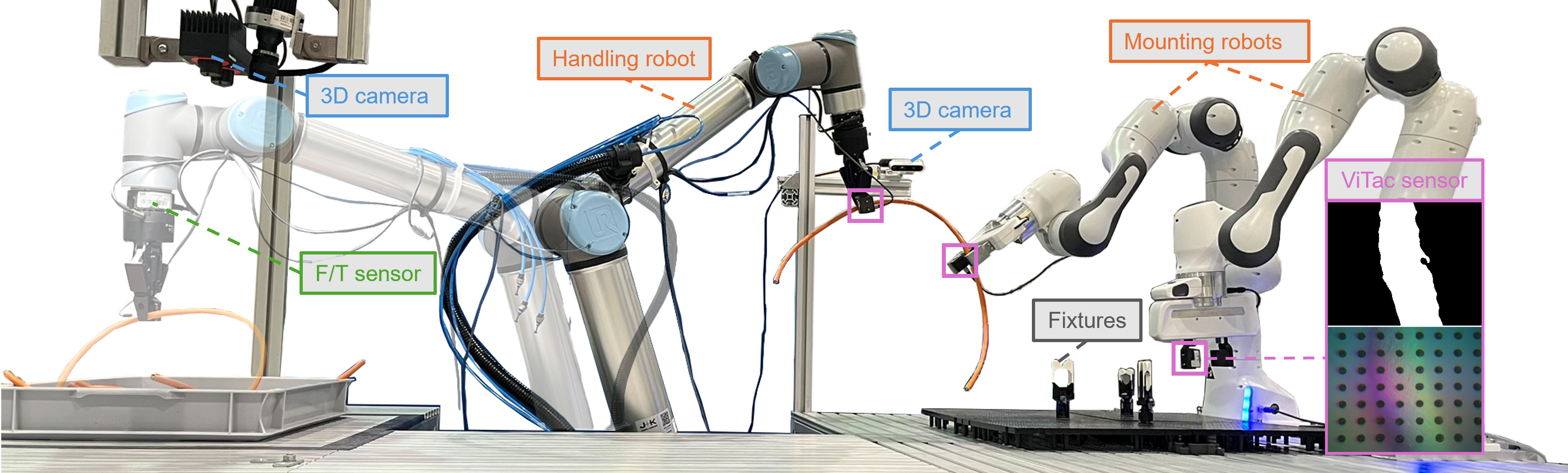}\\
        \captionof{figure}{Overview of the assembly process. The handling robot on the left picks one DLO from a bin full of DLO instances and hands it to one of the mounting robots on the right. The two mounting robots then collaboratively mount the DLO onto designated fixtures. The DLO's status is monitored by RGB-D cameras, F/T and ViTac sensors throughout the process.}
        \label{fig:setup}
    \end{center}
    \vspace{4pt}
    }]
}
\begin{document}

\maketitle 


\begin{abstract}
Industrial assembly of deformable linear objects (DLOs) such as cables offers great potential for many industries. 
However, DLOs pose several challenges for robot-based automation due to the inherent complexity of deformation and, consequentially, the difficulties in anticipating the behavior of DLOs in dynamic situations. 
Although existing studies have addressed isolated subproblems like shape tracking, grasping, and shape control, there has been limited exploration of integrated workflows that combine these individual processes.

To address this gap, we propose an object-centric perception and planning framework to achieve a comprehensive DLO assembly process throughout the industrial value chain. 
The framework utilizes visual and tactile information to track the DLO's shape as well as contact state across different stages, which facilitates effective planning of robot actions. 
Our approach encompasses robot-based bin picking of DLOs from cluttered environments, followed by a coordinated handover to two additional robots that mount the DLOs onto designated fixtures.
Real-world experiments employing a setup with multiple robots demonstrate the effectiveness of the approach and its relevance to industrial scenarios.

\end{abstract}

\section{Introduction}
The manipulation of deformable linear objects (DLOs), such as cables, wires, or hoses, is a prevalent yet insufficiently automated process in industrial settings.
The high number of degrees of freedom (DoF) and complex physical properties of DLOs result in time-varying configurations, making their behavior in dynamic scenarios difficult to model accurately. 
Consequently, both the perception and manipulation of DLOs present persistent challenges in robotic automation~\cite{zhu2022challenges}. 
These challenges lead to a low degree of automation in DLO-related tasks, necessitating substantial parts of manual labor even within highly automated production systems~\cite{OLBRICH2022653}.

Recent advancements in robotics research introduced various methodologies to address specific subproblems such as shape tracking~\cite{Caporali.2023, xiang2023trackdlo}, accurate grasping~\cite{dirr2023_2, zhang2024}, and shape control ~\cite{zhu2018dual, suberkrub@feel, yu2023generalizable} of DLOs. 
Despite these developments in isolated tasks, limited attention has been dedicated to addressing the overall assembly process~\cite{SALUNKHE2023696}. 

Assembly is inherently a long-horizon, multi-step operation comprising singulation, handling, and precise mounting of DLOs onto designated fixtures or components~\cite{makris2023automated}. 
The distinct nature of each stage in assembly introduces heterogeneous challenges, making holistic planning and execution particularly intricate.
For instance, assembly frequently occurs in cluttered environments where the target DLO interacts with contextual objects, such as fixtures or other DLO instances ~\cite{chen2024real}. 
These interactions increase the complexity of visual shape tracking and introduce the additional challenge of accurately monitoring DLO's contact with the environment~\cite{Chang2022ManipulationOD}.
Furthermore, large spatial transfers of the DLOs and dynamic scenarios throughout the assembly process make it difficult for a single sensor to track the DLO throughout the process and thus necessitate multi-modal perception. 
Beyond perception, effective assembly requires coordinated planning among multiple robots to ensure seamless execution. 
Additionally, solutions for individual stages have to be aligned regarding general assumptions and boundary conditions, such as compatible grasp poses on the DLO's global 3D shape, to ensure a robust workflow. 
Despite potential challenges, automating the assembly process carries significant implications across multiple fields, from industrial manufacturing to construction.


    

In this work, we introduce a holistic, object-centric perception and planning framework for DLO assembly that bridges previously isolated tasks into an integrated pipeline. 
As is shown in Fig.~\ref{fig:setup}, the comprehensive pipeline consists of robot-based bin picking from cluttered bins (singulation), coordinated inter-robot transfer, and mounting of the DLO onto predefined fixtures.
In particular, to address the complexities resulting from interconnected stages, we utilize multi-modal perception by combining visual, tactile, and force information to continuously monitor the object's state.
Coordination among multiple robots is planned accordingly to adapt to unexpected movements and deformations during DLO assembly in cluttered and dynamic industrial environments.
This study aims to provide insights into previously overlooked challenges associated with integrating distinct stages of DLO manipulation and to offer a practical approach toward improved efficiency and adaptability in automated, robotic DLO assembly.

\section{Related Work}~\label{sec: related}
\vspace{-2em}
\subsection{Visual DLO Perception}
The visual perception of DLOs remains particularly challenging due to their absence of distinct features and the dynamic variation in appearance, a consequence of their flexibility~\cite{Caporali.2023}. Additionally, their typically small diameters complicate perception using 3D sensors~\cite{Cop.2021}. 

Several deep learning-based approaches have been proposed for the 2D segmentation of DLOs, primarily utilizing data-driven methodologies for semantic segmentation~\cite{Caporali.2023, choi2023mbest}. 
Subsequently, various tracing and merging techniques are applied to delineate individual DLO instances for instance segmentation~\cite{Caporali.2023, choi2023mbest}. 
However, all methods mentioned consider at most ten DLO instances depicting few intersections and little occlusion or self-intersection of a singular object. 
Weaker performances are observed in case of high occlusion. 

Furthermore, deep learning-based approaches require extensive, annotated datasets for effective training, which is especially hard for DLOs due to their infinite possible configurations. 
The emergence of zero-shot foundation models, 
such as Segment-Anything Model (SAM)~\cite{kirillov2023segment} and Segment-Anything Model 2 (SAM2)~\cite{ravi2024sam}, presents a paradigm shift for DLO perception~\cite{zhang2024}. 
These models enable segmentation without the need for training on specific datasets, eliminating the dependency on extensive labeled data.  
There are a few approaches that consider segmentation for DLOs with SAM or SAM2. 
ISCUTE~\cite{kozlovsky2024iscute} uses SAM to refine the initial mask generated by CLIPSeg~\cite{luddecke2022image}, a different model for zero-shot segmentation, for improved accuracy. 
Zhaole et al.~\cite{zhaole2023robust} employ SAM using a text prompt to reconstruct highly occluded DLOs in 3D. 
However, ISCUTE only considers sparse scenes, and Zhaole et al. consider occlusion by objects with differing appearances. 

This work proposes a pipeline to utilize SAM2 to generate a functional instance segmentation of DLOs in complex and cluttered scenes. 
The proposed approach operates without requiring training data, relying on automatically sampled input points without human intervention.

\subsection{Multi-modal DLO Perception}
While vision is the dominant information source for DLO perception, there has been a growing interest in incorporating force and tactile sensory data for DLO perception.
Force/torque sensors (F/T sensor) integrated in joints or mounted on wrists have been proven beneficial for estimating DLO's contact status with environmental objects, especially fixtures. 
Suberkrub et al. introduce a method that detects contact points on DLOs with force-torque information only by leveraging the tension~\cite{suberkrub@feel}. 
Building on this, recent works extend the use of force measurements to estimate contact establishment and changes~\cite{chen@contact, chen2024real} or to maintain tension during manipulation~\cite{zhang2024harnessing}.
In this work, we use the force information to detect and handle contact with the environment, especially in solving the entanglements between DLOs.

Tactile sensors mounted on gripper tips offer further insights into the local contact area when the robot is in contact with the DLO.
This allows extracting the shape of the grasped DLO portion, which is otherwise occluded from vision.
Such information has been used to reconstruct the DLO's shape for contour following~\cite{monguzzi2023tactile, yu2024hand} and subsequent DLO routing~\cite{monguzzi2024potential}.
Additionally, vision-based tactile sensors (ViTac sensors) also provide a rough estimation of contact forces.
Wilson et al. leveraged such function to develop tactile-driven motion primitives to achieve DLO routing~\cite{wilson2023cable}.

In this work, we aim to integrate local tactile shape information with the global visual estimation for more accurate tracking of the DLO's 3D shape during manipulation and under occlusions.
The most comparable work to ours is Caporali's~\cite{caporali2021combining}, which relies on a 2D camera image for the initial grasp, and uses a local shape tactile image to evaluate and adjust the grasping pose.
Our works differ from theirs in that we also use tactile correction to refine global shape tracking from 3D camera information, which aims to enhance the motion planning of another robot for collaboration.

\subsection{DLO Manipulation}  
The physical properties of DLOs introduce additional complexities for manipulation tasks such as bin picking~\cite{Zhang_2023, grard2020} or shape control~\cite{yu2023generalizable, zhu2022challenges}. 
The high degree of flexibility in DLOs results in trajectory variations that are hard to predict, making detection and segmentation complicated.
Additionally, dense scenes in industrial scenarios lead to occlusions and physical entanglements, raising the difficulty of grasp planning and object retrieval~\cite{Zhang_2023}.

Despite advances in robot-based bin picking of rigid objects, the industrial-grade automation of bin picking for DLOs remains an open challenge~\cite{dirr2023_2}. 
Dirr et al.~\cite{Dirr.2024} propose an instance-aware bin picking approach for DLOs designed for structured environments but do not address the complexities of cluttered scenes (object-oriented). 
Conversely, Zhang et al.~\cite{Zhang_2023} aim to find a grasping pose suitable for the gripper in use (gripper-oriented). 
Whereas the second approach considers cluttered scenes, the lack of instance awareness gives little control over the position of the grasping pose on the DLO's global 3D shape.

\setcounter{figure}{1}
\begin{figure*}[ht!]
    \centering
    \includegraphics[width=0.95\textwidth]{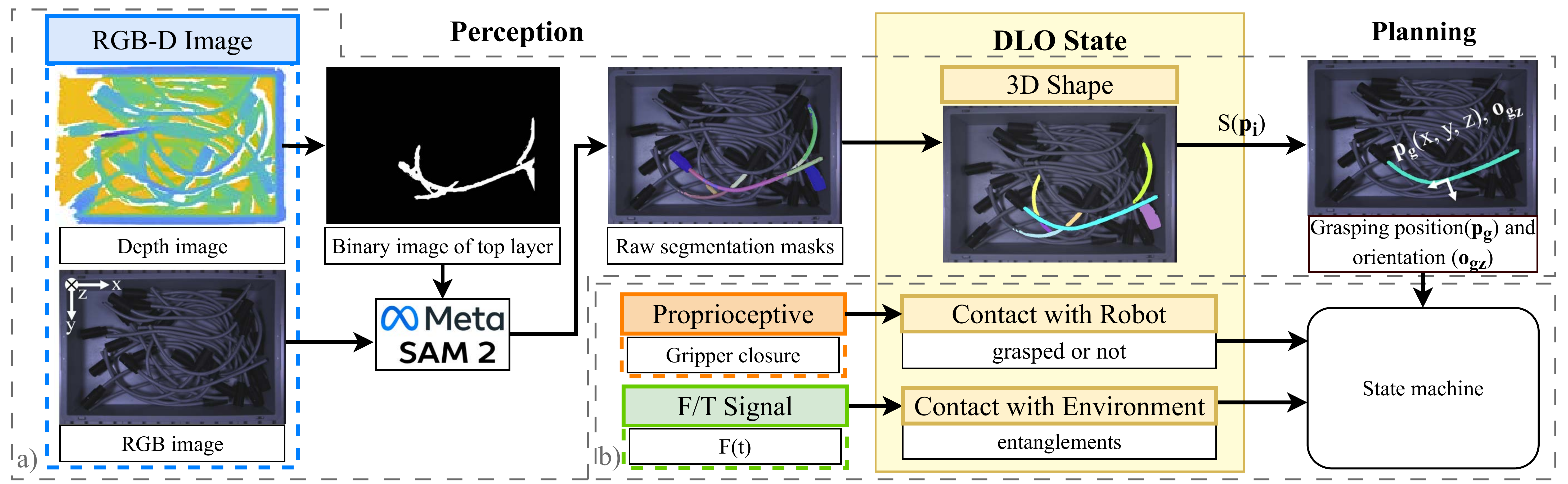}
     \caption{Bin picking pipeline. a) The 3D shape of the selected DLO is extracted from the segmentation of the top layer of the bin. A grasping position, $\bf{p_{g}}$, along with the gripper's orientation around the z-axis, $\bf{{o}_{g_z}}$ (visualized by the arrows), is determined accordingly. b) The picking motions are planned using a state machine that integrates the 3D shape as well as the contact status obtained from multi-modal perception.}
    \label{fig:method_BP}
    \vspace{-1.5em}
\end{figure*}

This work proposes a model-oriented approach for bin picking of DLOs in cluttered scenes. 
Since model-oriented methods require near-perfect instance segmentation, which is challenging under high occlusion, we propose a partial instance segmentation approach that ensures a grasping pose suitable for both the gripper and the object.
In combination with the shape control approach proposed in our previous work~\cite{chen@contact}, the joint framework manages to achieve a comprehensive workflow for DLO assembly from bin picking to mounting.



\section{Methodology}
We define the workflow for robot-based DLO assembly in manufacturing as a sequence of the following stages:

\begin{enumerate}
    \item Singulation: Given an initial supply of multiple DLOs, the first step is to select a single DLO for assembly. 
    In our framework, a material-handling robot arm selects a DLO from a bin containing multiple DLOs, ensuring it is properly separated.
    \item Handling: Once picked up, the DLO is transported to other robotic systems for further manipulation, constituting a standard material handling step in manufacturing. 
    In our framework, the handling robot hands it over to one of the mounting robots.
    \item Mounting: In the final stage, the DLO is aligned and secured onto predefined fixtures. 
    Our framework employs two collaborative robotic arms to precisely position and fasten the DLO into place.
\end{enumerate}

To ensure the smooth execution of each stage and their seamless integration, the state of the DLO must be continuously monitored throughout the assembly process. 
Specifically, the state representation includes the following key aspects:
\begin{itemize}
    \item Global 3D shape. The DLO’s shape is represented as a sequence of 3D points $S = (\mathbf{p_1}, \mathbf{p_2}, ...\mathbf{p_N})$, where $\mathbf{p_i}=(x_i, y_i, z_i)$ represents a 3D position. 
    The unit tangent pointing from $\mathbf{p_i}$ to $\mathbf{p_{i+1}}$ is defined as $\mathbf{t_i} = \frac{\mathbf{p_{i+1}} - \mathbf{p_{i-1}}}{|\mathbf{p_{i+1}} - \mathbf{p_{i-1}}|}$.
    Accurate perception of the current 3D configuration of the DLO is crucial for both the initial picking phase and subsequent manipulation processes.
    \item Contact with the robot. This state element determines whether the DLO has been successfully grasped by a robot. 
    A reliable attachment ensures the DLO is secured and ready for further manipulation.
    \item Contact with the environment. 
     This includes the DLO's interactions with the contextual objects, such as entanglement with other DLO instances during supply, which affects singulation of the DLO, as well as contact with obstacles, such as the bin itself or fixtures during mounting, which influences the precision and robustness of the whole assembly process.
\end{itemize}

In this section, we will elaborate on how we extract and leverage the aforementioned state information for object-centric planning in each stage of the assembly process.



\subsection{Singulation}
\label{subs:binp}
To replicate industrial conditions, bin picking is performed on a densely packed set of DLOs within a container, resulting in a cluttered, occluded scene.
Since bin picking serves as an upstream process, the grasping pose on the DLO must accommodate constraints imposed by subsequent operations, requiring instance-level awareness of the selected DLO.
To address this, an object-centric approach is adopted, drawing on prior work~\cite{Dirr.2024}.  
In highly occluded environments, instance segmentation remains challenging. 
Thus, the proposed approach prioritizes functional instance segmentation and selecting suitable grasping points on visible DLO parts. 

\textbf{Local Visual Perception in Cluttered Scenes}
\label{sec:img_acq}
DLOs stored in bulk are prone to entanglement, which makes the downstream handling complicated. 
Therefore, the goal is to identify a suitable DLO for picking and to extract relevant information about its 3D shape.
DLOs in the topmost layer of the pile are assumed to be less likely entangled or occluded, making them preferable for picking \cite{inagaki2019detecting}. 
Among these, the DLO with the longest visible segment is considered the most suitable.
To extract relevant information, the foundation model SAM2 is utilized for segmentation, using a top view RGB-D image of the bin as input (Fig.~\ref{fig:method_BP} (a)). 
The subsequent method refers to the coordinate system depicted in the figure.

Since only DLOs in the top layer of the bin are relevant, the segmentation process focuses on this layer.
As the $z$-coordinate of the DLO determines whether it should be segmented, the depth data of the scene serves as input for SAM2. 
To leverage geometric information, the ``point prompt" function of SAM2 is used, in which several input points are utilized to ensure robustness and mitigate the model’s inherent non-determinism. 
Each point prompt results in an individual segmentation mask for the scene.
The generated segmentation masks are subsequently merged through post-processing.

To extract the depth data for the top layer, a threshold-based search is applied to the depth image.
Once a predefined area $A_{threshold}$ of contiguous pixels is reached within the depth image, the binary image of the top layer is extracted and skeletonized, yielding central paths that approximate the structure of the most accessible DLOs.
From these paths, $N$ equidistant points are sampled as input prompts for SAM2, ensuring segmentation focuses on the most viable DLOs for bin picking.
The parameter $N$ is predetermined and can be adjusted to optimize performance.

The original RGB image remains the reference for segmentation as DLOs extending beyond the top layer may still be visible and should be included for a more comprehensive shape representation. 
Based on the $N$ point prompts, SAM2 generates $N$ segmentation masks, each associated with a confidence score. To refine these masks, Algorithm~\ref{alg_sam} is employed. 
The algorithm begins by sorting the masks in descending order of confidence scores. 
Each mask is then sequentially compared with the remaining ones.
If the Intersection over Union (IoU) between two masks exceeds the predefined merging threshold $T_{merge}$, they are combined into a single mask. 
This merging process is repeated iteratively, ensuring that overlapping regions are consolidated. 
Following the merging stage, each refined mask is evaluated against the processed set $M_{proc}$.
A mask is incorporated in $M_{proc}$ only if its IoU with any previously processed mask remains below the discard threshold $T_{discard}$, effectively filtering out redundant or low-quality segmentations. 

\begin{algorithm}
\caption{Sort, merge, and discard raw segmentation masks}
\small{
\begin{algorithmic}[1]
\Require Raw segmentation masks $M_{raw}$ from SAM2, $T_{merge}$ and $T_{discard}$ are adjustable parameters
\Ensure Processed masks $M_{proc}$
\State Initialize $M_{proc} \gets \emptyset$
\State Sort $M_{raw}$ by confidence score in descending order
\For{$i \gets 1$ to $|M_{raw}|$}
    \State $mask_i \gets M_{raw}[i]$
    \For{$j \gets i + 1$ to $|M_{raw}|$}
        \State $mask_j \gets M_{raw}[j]$
        \If{$\text{IoU}(mask_i, mask_j) > T_{merge}$}
            \State Merge $mask_i$ and $mask_j$ into $mask_i$
        \EndIf
    \EndFor
\EndFor
\For{$mask \in M_{raw}$}
    \If{$\forall m \in M_{proc} : \text{IoU}(mask, m) \leq T_{discard}$}
        \State Add $mask$ to $M_{proc}$
    \EndIf
\EndFor
\end{algorithmic}
}
\label{alg_sam}
\end{algorithm}

As is illustrated in Fig.~\ref{fig:method_BP}, the final set of processed masks corresponds to the least occluded DLOs in the top layer. These masks are skeletonized and pruned to extract $S(\bf{p_i})$, which serves as the basis for determining the grasping position $\mathbf{p_{g}}$ and the grasping orientation $\bf{o_g = (o_{g_x}, o_{g_y}, o_{g_z})}$.
$\bf{p_{g}}$ is calculated along the longest continuous skeleton, with its position relative to the skeleton length controlled by parameter $R$ to ensure proximity to one end of the DLO. 
As the desired picking pose is orthogonal to the bin, the rotation around the $X$-axis and $Y$-axis $\bf{o}_{g_x}$ and $\bf{o}_{g_y}$ are assumed to be zero.
The rotation around the $Z$-axis $\bf{{o}_{g_z}}$ is determined by the pitch angle of the 3D shape, i.e. the projection of tangent vector at the grasping point $\bf{t_g}$ on the $X$-$Y$ plane.
$\bf{p_{g}}$ and $\bf{{o}_{g_z}}$ then serve as inputs to the subsequent motion planning.

\begin{figure}[h]
    \centering  
    \includegraphics[width=0.95\columnwidth]{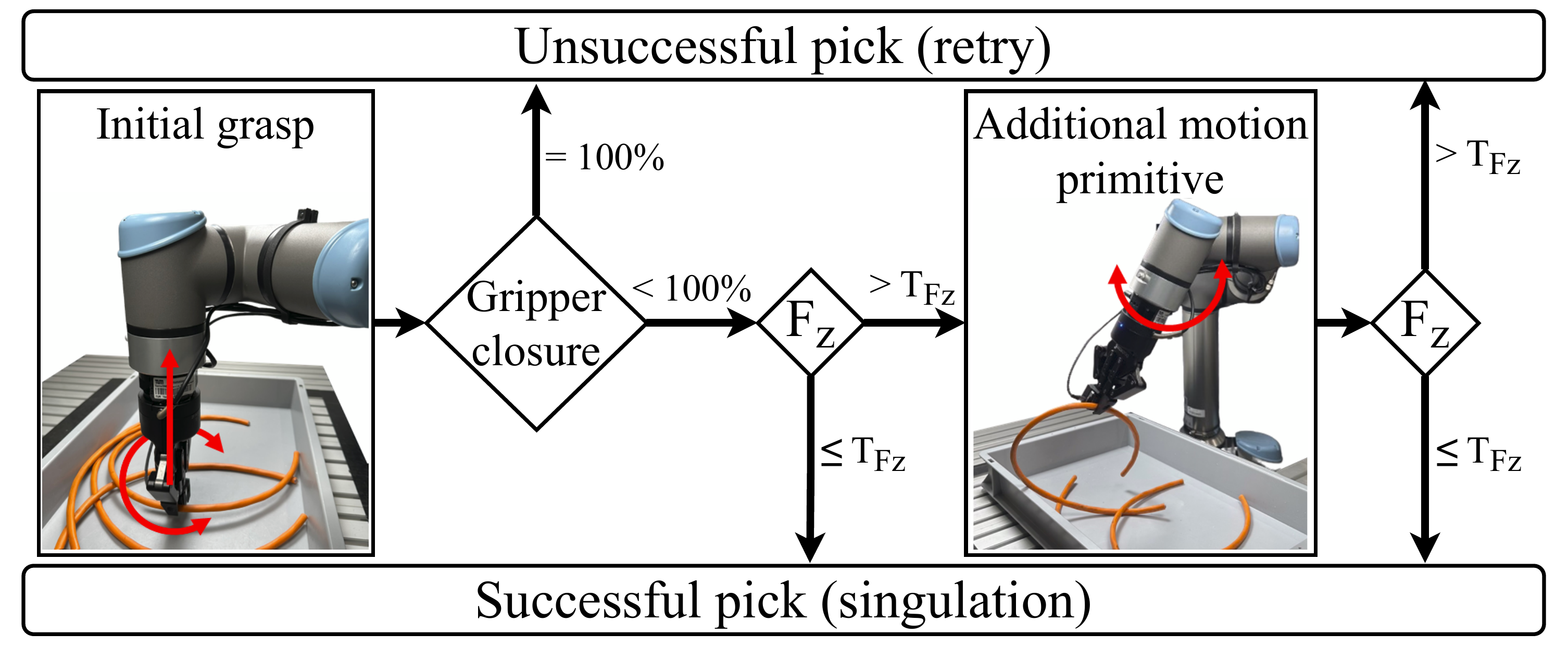}
    \caption{A state-machine-based approach forms the foundation for motion planning by leveraging proprioceptive and tactile data. Based on the perception, the robot will either execute an additional motion primitive or abort the pick in the event of anomalies.}
    \label{fig:method_MP}
\end{figure}

\textbf{Motion Planning}
 Once $\bf{p_{g}}$ and $\bf{{o}_{g_z}}$ are determined, the robot proceeds to pick the DLO from the bulk using multi-modal information as depicted in Fig.~\ref{fig:method_BP} (b). 
 To handle errors during manipulation, we propose a state machine-based approach (Fig.~\ref{fig:method_MP}).  
 Initially, the robot moves to $\bf{p_{g}}$ rotated according to $\bf{{o}_{g_z}}$ and closes its two-finger gripper. 
 Grasp success, and thus sufficient contact between DLO and the robot is inferred from the gripper closure. 
 If the jaws are completely closed, the attempt is deemed unsuccessful, prompting the robot to abort the pick and return to its home position. 
 Conversely, if the jaws remain partially open, the DLO is assumed to be grasped and lifted from the pile. 
To overcome the potential risk of entanglement, specific actions are designed to separate the DLO from its surroundings (Fig.~\ref{fig:method_MP}). 

Following the initial grasp, the contact status of the DLO is assessed using F/T data. 
To minimize noise from dynamic movements, the robot moves to a predefined static position and measures force along the z-axis.
Contact with a contextual object is inferred if the detected force exceeds the threshold $T_{F_z}$, calculated as:
\begin{equation}
T_{F_z} = m_{DLO} \cdot g \cdot \eta_{F_z}\text{,}
\end{equation}
where $m_{DLO}$ is the mass of the DLO, $g$ is the gravitational acceleration, and $\eta_{F_z}$ is a safety factor. 
In such cases, the robot performs additional  operations, such as a pendular motion (Fig.~\ref{fig:method_MP}), to dislodge entangled DLOs.

Once a DLO is successfully isolated, it is retained for handover to the robotic system responsible for the mounting process.

\begin{figure*}[t]
    \centering
    \includegraphics[width=0.95\textwidth]{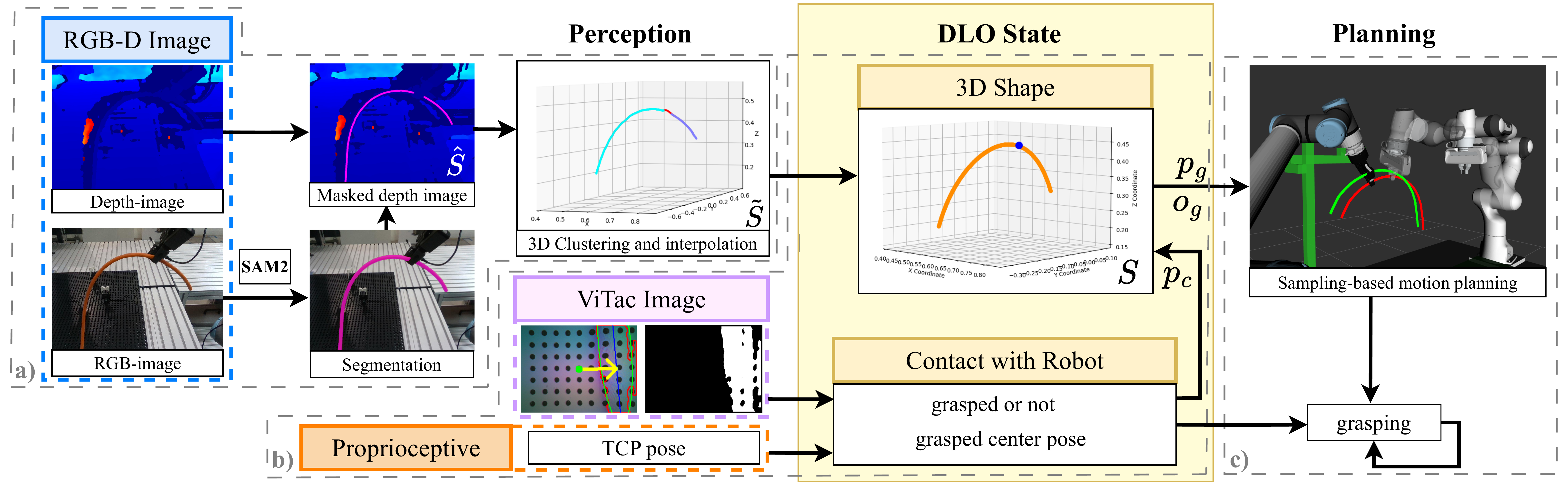}
    \caption{Handover pipeline. a) After extraction from segmentation, the DLO's raw 3D points $\hat{S}$ are clustered by DBSCAN (blue and purple), and the missing part is interpolated (red) to form an initial 3D model $\tilde{S}$. b) The grasping status as well as the grasping center are derived from the in-hand ViTac images and the robot's TCP pose. The yellow arrow on the ViTac image points from TCP to the grasping center. The final 3D shape $S$ (orange) is corrected with grasp center point (dark blue ball). c) Based on the corrected shape (red), the grasping for the second robot is planned.}
    \label{fig:method_handover}
    \vspace{-1.5em}
\end{figure*}

\subsection{Handover}\label{subsec:handover}
The handover operation necessitates precise monitoring of the DLO’s 3D shape, along with motion planning accordingly to determine an appropriate grasping pose as well as a collision-free trajectory.
Toward accurate online shape tracking, we align the global visual estimation to the local ground truth from proprioceptive and ViTac observations.

\textbf{Global Visual Estimation}
Primarily, an RGB-D camera tracks the DLO's 3D shape using a similar pipeline as depicted in Section~\ref{subs:binp}, resulting in a preliminary set of 3D points $\hat{S}$ (see Fig.~\ref{fig:method_handover} (a)).
The difference lies in that instead of top images, the camera now captures perspective observations, which has a broader range of view but also results in more cluttered scenarios and noisy depth data.
To address these challenges, we apply DBSCAN~\cite{ester1996density} to identify clusters and remove outliers in the depth data. 

For each point $\mathbf{\hat{p}}_i$, DBSCAN collects its $\epsilon$-neighborhood $N_\epsilon(\mathbf{\hat{p}}_i)$.
If the neighborhood contains a higher number of points than a threshold $T_n$, $\mathbf{\hat{p}}_i$ is classified as a cluster; otherwise, it is labeled as noise.  
Through this process, DBSCAN yields clusters $\{\hat{c}_i\}$, each representing a denoised, unoccluded DLO segment.
The clusters are then ordered based on their projection onto the DLO's principle axis that is computed through principle component analysis (PCA).
Subsequently, we apply regression methods to interpolate the missing portions and reconstruct the entire 3D shape $\tilde{S}$.

\textbf{Local Correction}
While visual perception provides an overview of the DLO’s shape, the extracted world coordinates are often noisy and prone to drift.
By contrast, when a robot grasps the DLO, the ground-truth pose of the grasped segment can be acquired directly through proprioceptive and tactile sensing. 
To enhance overall accuracy, we fuse this precise local measurement with the global visual estimation.

\begin{algorithm}[ht!]
\caption{Visual-tactile tracking}\label{alg: visual_estimate}
\small{
\begin{algorithmic}[1]
\Require Raw 3D shape $\hat{S}$ segmented by SAM2, in-hand grasping center $^{tcp}\mathbf{p}_g$, TCP pose $(\mathbf{p}_{tcp}, \mathbf{o}_{tcp})$
\Ensure Corrected 3D shape $S$
\State Unoccluded clusters $\{c_i\} = \mathrm{DBSCAN}(\epsilon, D_n, \hat{S})$
\State Principle axis $\mathbf{v}_{pca} = \mathrm{PCA}(\{c_i\})$
\State Ordered clusters $(c_1, c_2, ..., c_N) = \mathrm{Sort}(\{c_i\}, \mathbf{v}_{pca})$
\State Initialize $\tilde{S} \gets \emptyset$
\For{$i \gets 1$ to $N-1$}
    \State $\tilde{S} \gets (\tilde{S}, c_i)$
    \State$\tilde{s_i} = \mathrm{PolynomialFitting}(c_i, c_{i+1})$
    \State $\tilde{S} \gets (\tilde{S}, \tilde{s_i})$
\EndFor
\State Visual estimated shape $\tilde{S} \gets (\tilde{S}, \tilde{s_N})$
\State Grasping center $\mathbf{p}_c = \mathbf{o}_{tcp} \cdot ^{tcp}\mathbf{p}_g + \mathbf{p}_{tcp}$
\State Corrected shape $S \gets$ Equ.~\ref{eq: correct}
\end{algorithmic}
}
\end{algorithm}

Once the DLO is grasped by a robot equipped with a ViTac sensor, the contact image provides access to its in-hand position $^{tcp}\mathbf{p}_g$ in the frame of tool center point (TCP) (see Fig.~\ref{fig:method_handover} (b)).
Combined with the robot's TCP pose obtained from proprioceptive information, the center position of the grasped DLO segment in the base frame $\mathbf{p}_c$ can be computed.
Since this grasping point is occluded from the initial visual observation $\hat{S}$, we identify its corresponding point on the DLO by selecting the nearest cluster center among the missing segments after DBSCAN filtering, denoted as$\tilde{\mathbf{p}_{c}} \in \tilde{S}$.
The estimated shape $\tilde{S}$ is then aligned with the local measurement by translation to ensure that $\tilde{\mathbf{x}}_{c}$ coincides with $\mathbf{p}_c$:
\begin{equation}\label{eq: correct}
S  = \mathbf{T}_{\text{correct}} \odot \tilde{S}, \ 
\mathbf{T}_{\text{correct}} = \begin{bmatrix}
    \mathbf{I}_{3\times 3} & \mathbf{p}_c - \tilde{\mathbf{p}}_{c} \\
    \mathbf{0}^T & 1
\end{bmatrix}\text{.}
\end{equation}
The corrected shape $S$ serves as the final tracking result for subsequent robotic motion planning. 
The entire tracking procedure is summarized in Algorithm~\ref{alg: visual_estimate}.

\textbf{Motion Planning}
Based on the final tracking result (red curve in Fig.~\ref{fig:method_handover} (c)) corrected from original visual estimation (green curve), we plan the motion of the second robot to co-grasp the DLO.
Similar to grasping in Section~\ref{subs:binp}, the grasping position $\mathbf{p}_{g}$ is set at a certain offset $L_{g}$ from the first robot's grasping point.
Grasping orientation $\mathbf{o}_g$ is selected to align with the DLO's local shape at the grasping point:
\begin{equation}
    \mathbf{o}_{g_x} = \mathbf{t_g}, \ 
    \mathbf{o}_{g_y} = \mathbf{n_g}, \ 
    \mathbf{o}_{g_z} = \mathbf{o}_{g_x} \times \mathbf{o}_{g_y}\text{,}
\end{equation}
where $\mathbf{t_g}$ and $\mathbf{n_g}$ are tangent and normal vectors of the DLO at $\mathbf{p}_g$.
For each grasping pose, we solve inverse kinematics to obtain robot joint configurations and employ a sampling-based planner to find collision-free trajectories.
The optimal path is selected among the feasible ones, by minimizing the joint path cost.

After executing the planned trajectory, the second robot's TCP may deviate slightly from $\mathbf{p}_g$ due to errors accumulated from e.g., simulation modeling or robot motion control.
Therefore, a local grasping correction is applied to further align the grasping point to the TCP.
As depicted by the yellow arrow on ViTac images in Fig.~\ref{fig:method_handover} (b), the robot will move for a small displacement pointing current grasping point to the TCP, and retry grasping.
This process is repeated until the desired alignment for subsequent mounting motion is achieved.


\subsection{Mounting}
After receiving the DLO from the handling robot, the mounting robots secure it onto a predefined set of clip fixtures.
For the mounting task, we employ the dual collaborative manipulation framework for deformable linear objects developed in our prior work~\cite{chen@contact}. 
One of the mounting robots moves the DLO toward each designated fixture.
Upon reaching a fixture, the second mounting robot joins the process of inserting the DLO into the clip fixture.
The joint trajectory of both robots, from their current poses to the fixture, is also planned by sampling-based motion planner to ensure collision avoidance between the robots and environmental objects.
During the fixing process, both robots apply a feedforward force to securely push the co-grasped DLO segment into the clip. 
This approach-and-fix routine is iteratively executed for each fixture until the DLO is mounted onto all designated fixtures.

\section{Experiments}
Since the approach presented in this paper introduces advancements in both individual stages and the overall assembly process,  experimental evaluation is divided into sub-experiments for bin picking and the handover task. 
Finally, the complete assembly pipeline is tested.

The experimental setup, illustrated in Figure~\ref{fig:setup}, consists of a UR10e robotic manipulator as the handling robot alongside two Franka Emika Panda robots as mounting robots. 
A Roboception rc\_visard 65c stereo camera, with a RandomDot pattern projector, is positioned 80 cm above the UR10e’s workspace for perception in bin picking.
A Bota Systems SensONE 6 DoF F/T-sensor is integrated into the UR10e, as preliminary evaluations indicated that the robot’s built-in sensor lacked the required precision.
A Robotiq 2F-85 gripper is used for the picking experiments.
Additionally, each robotic system is equipped with a GelSight Mini ViTac tactile sensor, embedded in one of the gripper jaws.
To oversee the handover process, an Intel RealSense D435 depth camera is strategically mounted at an inclined angle above the workspace.

\subsection{Bin Picking}

\begin{figure}[t!]
    \centering
    \includegraphics[width=0.9\columnwidth]{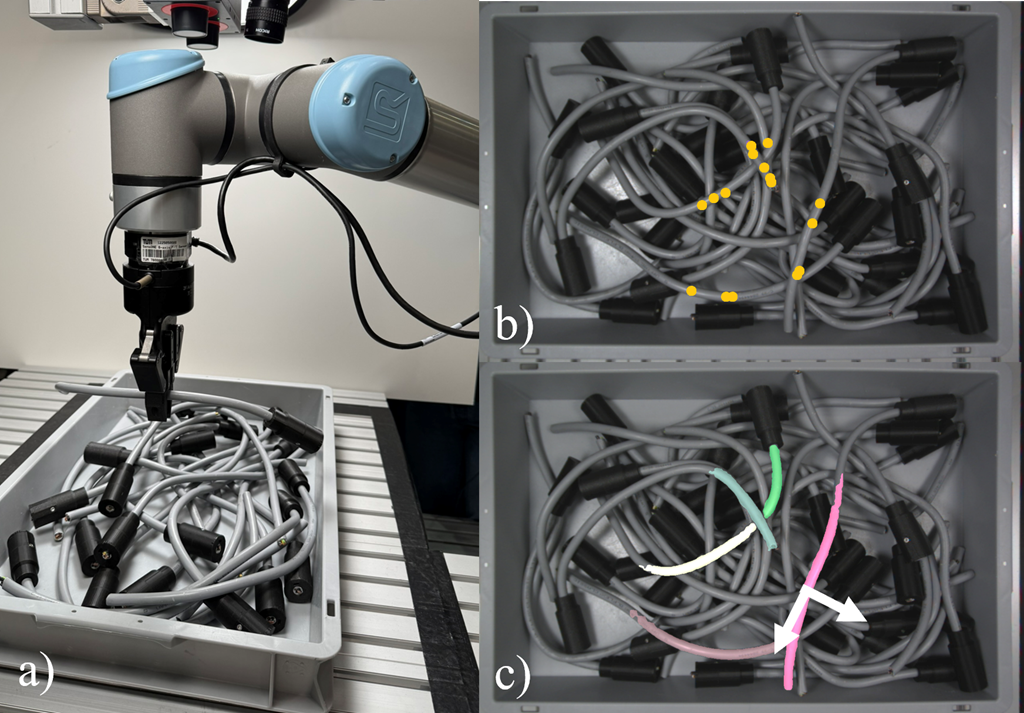}
    \caption{a) Experimental setup for robot-based bin picking, b) Point prompts for SAM2 generated from the depth data, c) Cluttered DLO scene in a bin with segmentation masks, $\bf{p_g}$ and $\bf{o_{g_z}}$, Parameter values, selected based on preliminary tests, are as follows: $\eta_{F_z} = 1.175$, $N = 20$, $T_{merge} = 40\%$, $T_{discard} = 10\%$, and $A_{threshold} = 15000$ pixels. With $R = 0.5$ $\bf{p_{g}}$ is derived closer to the center of gravity to compensate for weight distribution changes due to connectors. }
    \label{fig:setup_bin_picking}
    \vspace{-1.5em}
\end{figure}

The bin picking experiments utilize the setup illustrated in Figure \ref{fig:setup_bin_picking}. 
Since the ViTac sensor is not used for bin picking and obstructs grasping due to its bulky design, custom 3D-printed gripper jaws are employed in its place for the separate evaluation of bin picking.

The experiment involves removing 31 DLOs from a load carrier fixed to a workbench. Each DLO, modeled after high-voltage cables used in electric vehicle manufacturing, has an 11 mm diameter, weighs 0.13 kg, and is fitted with 3D-printed dummy connectors to induce entanglements, simulating industrial conditions.

A top-view RGB-D camera captures the bin, and images are processed following the methodology in Section \ref{subs:binp} to generate a grasp pose. 
The robot then executes the pick operation using motion primitives guided by sensor feedback.
To assess pick success, the robot places each DLO at a predefined table position. 
The process iterates until all DLOs are picked or no further grasp poses can be generated. 

A pick is considered successful if the DLO is isolated from the pile and placed on the table. 
Failure occurs if the DLO is dropped, multiple DLOs are grasped simultaneously, or no valid grasp pose is generated after two attempts.


The method is evaluated based on the success rate, defined as the ratio of successful picks to total attempts. Table \ref{tab:pick_performance} presents the results.

\begin{table}[h]
    \centering
    \caption{Results of the bin picking experiments}
    \renewcommand{\arraystretch}{1.1} 
    \setlength{\tabcolsep}{2pt} 
    \resizebox{\columnwidth}{!}{ 
    \begin{tabular}{c c c c c c}
        \hline
        \textbf{Bin} & \textbf{Number} & \textbf{Successful} & \textbf{Errors} & \textbf{Entanglements} & \textbf{Success} \\
        & \textbf{of DLOs} & \textbf{Picks} & & \textbf{Detected} & \textbf{Rate [\%]} \\
        \hline
        1 & 31 & 27 & 4 & 7 & 87.1 \\
        2 & 31 & 28 & 3 & 6 & 90.3 \\
        3 & 31 & 28 & 3 & 4 & 90.3 \\
        4 & 31 & 28 & 3 & 5 & 90.3 \\
        5 & 31 & 28 & 3 & 7 & 90.3
        \\
        \hline
        Overall &155&139&16&27& 89.7\\
        \hline
    \end{tabular}
    } 
    \label{tab:pick_performance}
\end{table}

The system achieves an average success rate of 89.7\%, demonstrating high reliability in cluttered environments. The most common sources for failures were unfavorable positioning of the DLO in the gripper, the absence of a suitable grasping point, or the DLO getting lost during manipulation.

\subsection{Tracking and Handover}
We then conduct experiments to evaluate the corrected 3D shape tracking described in Section~\ref{subsec:handover} and its impact on the DLO handover process.

\begin{figure}[t!]
    \centering  
    \begin{tikzpicture}
        \node [inner sep=0pt] (russell) at (0,0)
        {\includegraphics[width=0.48\textwidth]{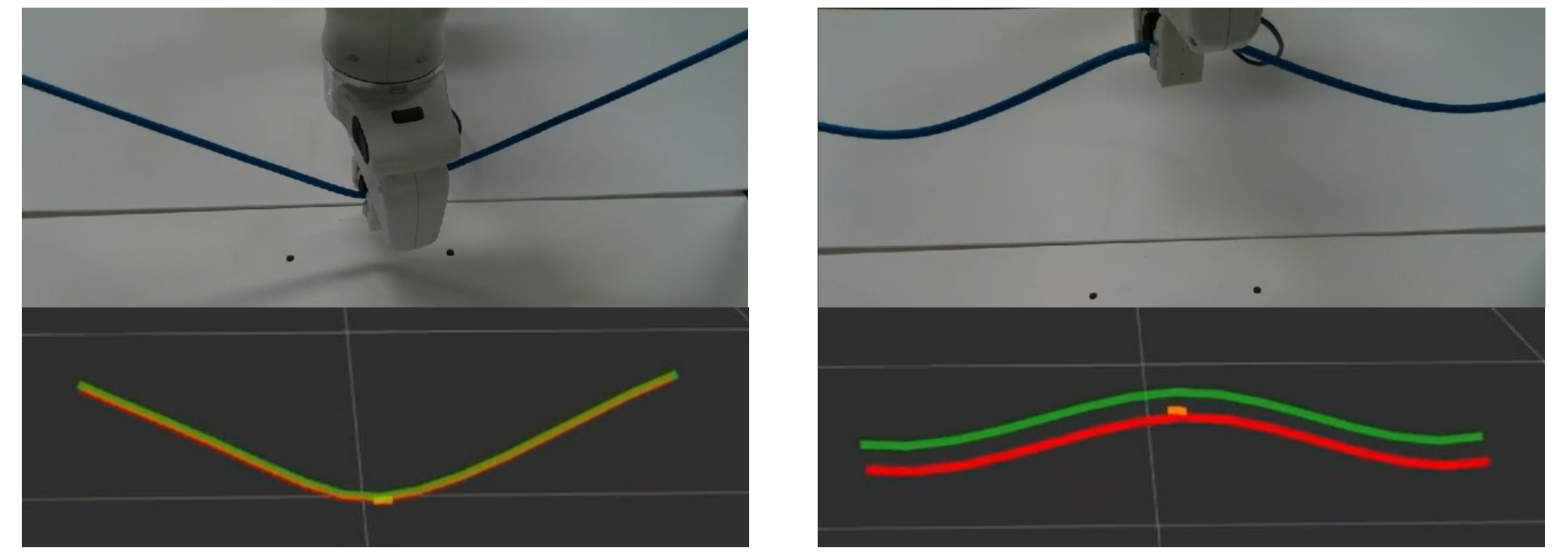}};
        \node [inner sep=0pt] (russell) at (0,-4.3)
        {\includegraphics[width=0.4\textwidth]{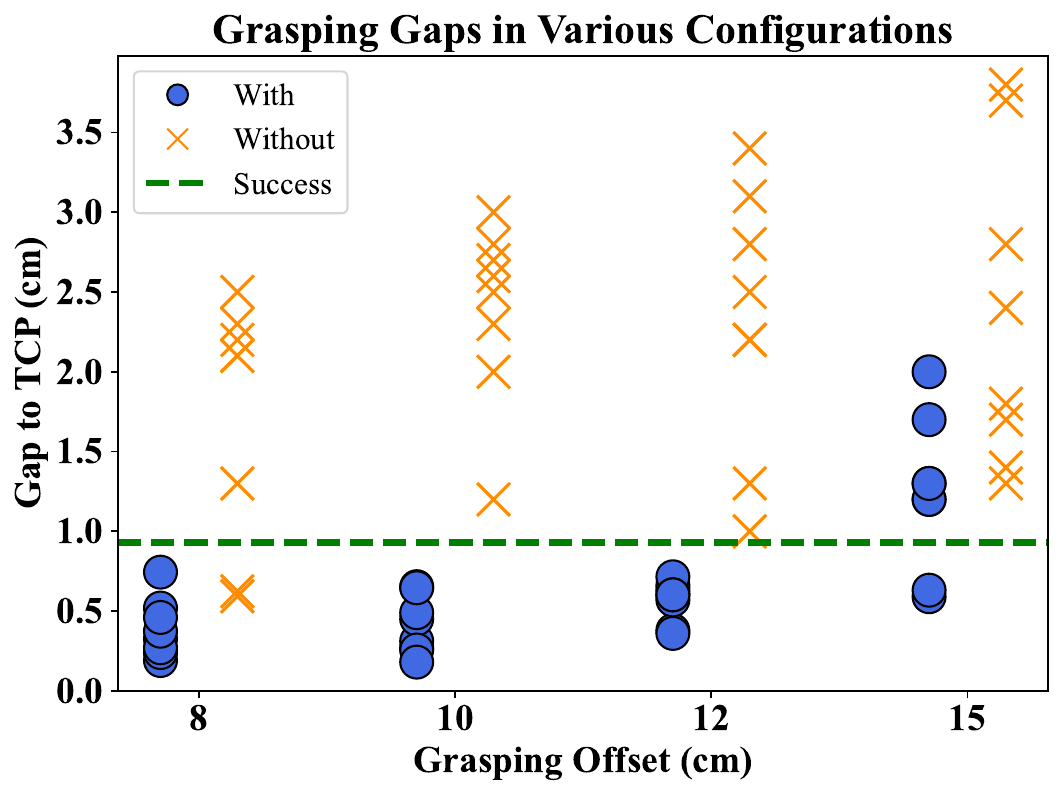}};
        \node[text=white] at (-3.95,-1.20) {a)};
        \node[text=white] at (0.4,-1.20) {b)};
        \node[] at (-3.95,-6.67) {c)};
    \end{tikzpicture}
    \caption{Handover experiments results. a) and b) Examples of small and large corrections. The raw visual estimation is marked in green, while the corrected model is marked in red. c) Gaps between the grasping point and the robot's TCP with (circle) and without (cross) local correction. Points below the green line are success.}
    \label{fig:handover_exp}
    \vspace{-1.2em}
\end{figure}

\begin{figure*}[t!]
    \centering  
    \begin{tikzpicture}[scale=0.9, every node/.style={scale=0.9}]
        \node[anchor=center] (img1) at (0, 0) 
        {\includegraphics[width=0.95\textwidth]{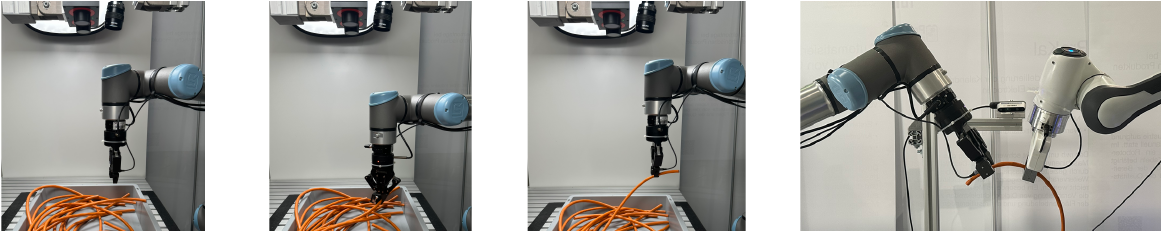}};
        
        \node[anchor=center] (img2) at (0, -4.02) 
        {\includegraphics[width=0.95\textwidth]{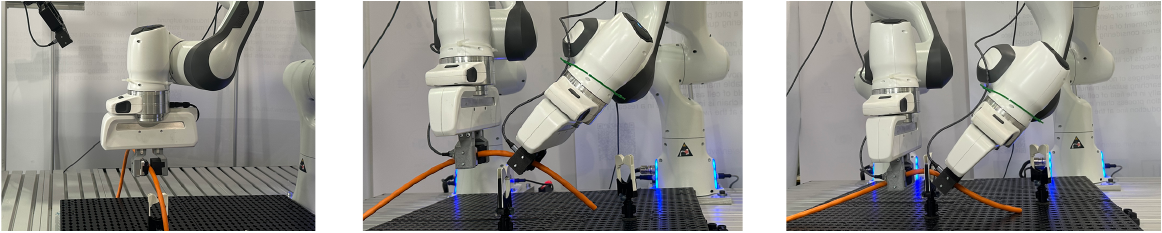}};   

        \node[anchor=center] (cap1) at (-6.9, -1.95) {Move to  $\mathbf{p}_g$};
        
        \node[anchor=center] (cap2) at (-3,-1.95) {Initial grasp};
        \node[anchor=center] (cap3) at (1.0, -1.95) {Singulation};
        
        \node[anchor=center] (cap4) at (5.72, -1.95) {Handover to the mounting robot};

        \draw[->, thick] (cap1.east) -- (cap2.west);
        \draw[->, thick] (cap2.east) -- (cap3.west);
        \draw[->, thick] (cap3.east) -- (cap4.west);
        \draw[->, thick] (cap4.east) -- (8.45,-1.95);

        \node (cap6) at (-5.98, -5.95) {Robot approaches fixture};
        \node (cap7) at (-0.17, -5.95) {Second robot supports};
        \node (cap8) at (5.68, -5.95) {Robots mount the DLO};

        \draw[->, thick] (-8.4, -5.95) -- (cap6.west);
        \draw[->, thick] (cap6.east) -- (cap7.west);     
        \draw[->, thick] (cap7.east) -- (cap8.west);

    \end{tikzpicture}
    \vspace{-2mm}
    \caption{The comprehensive assembly process performed by the three robots step-by-step. Changes in the parameter values for the bin picking method are as follows: $T_{merge} = 30\%$ and $A_{threshold} = 4000$ pixels. }
    \label{fig:exp_process}
    \vspace{-1em}
\end{figure*}

\textbf{Tracking Correction}
Due to the challenges in obtaining ground-truth 3D shapes of the DLO in real-world scenarios, we firstly evaluate the tracking performance, particularly the effectiveness of local correction, by comparing the reconstructed 3D model with and without correction.
Once the DLO is grasped, the robot performs random movements to induce deformation. To enhance the complexity of deformation, we use a more flexible DLO with lower stiffness in this experiment.

Across 10 trials, each consisting of 40 frames, the average adjustment before and after local correction is $2.34$ cm.
Cases of large or small offsets are depicted in Figure~\ref{fig:handover_exp} (a) and (b).
These results indicate that the local information does refine the reconstructed 3D shape, although the degree of correction may differ from position to position.
The average processing time for each frame is around $1.9$ seconds, allowing the tracking pipeline to operate online during manipulation.

\textbf{Handover results}
We further evaluate the accuracy of DLO handover between the handling and mounting robots.
To enable local correction, the UR10e is equipped with an in-hand ViTac sensor mounted on its finger.
After grasping, the UR10e will move the DLO to a certain position, waiting for one Panda robot to take it over.
The 3D shape is reconstructed using RGB-D images from the calibrated RealSense camera, the UR10e's proprioceptive information and the ViTac images. 
Upon initial grasping, the Panda has three chances to adjust its grasping position to align with its TCP with the assistance of ViTac sensors.

We ran handover experiments across four configurations, each involving variations in the UR10e's TCP position and the grasping point on the DLO.
In each configuration, we evaluate the grasping offset $L_g$ ranging from $8$ cm to $15$ cm. 
The resulting gaps between the grasping point and the Franka's TCP from 32 trials are recorded in Figure~\ref{fig:handover_exp} (c).
An handover with a gap below $0.93$ cm is considered successful, which corresponds to the field of view of the GelSight sensor. 
The results show that the local correction reduces the average gap by $1.54$ cm, and remarkably increase the handover success rate from $6.5\%$ to $81.25\%$.
Nevertheless, as the grasping offset grows, the handover accuracy may decrease, even when the correction is enabled.
This degradation is likely due to the grasping point shifting closer to the edge of camera view, where depth data experience greater distortion, and the local grasp information becomes less helpful.

\subsection{Overall evaluation}

Finally, we validate our framework within the continuous assembly process.
The experimental setup involves power cables with a length of $60$ cm, a diameter of $9.5$ mm and a weight of $0.1$ kg. 
$R$ is set to $0.9$ to force the UR's grasping point close to one end of the DLO, facilitating the handover.
$L_g$ is set to $1.0$ to ensure robust handover while avoiding collisions.

The comprehensive process is illustrated in Fig.~\ref{fig:exp_process}.
Initially, the DLO is retrieved from a bin containing 18 DLOs following the bin picking method proposed in Section~\ref{sec:img_acq}. 
Once extracted successfully, the UR10e transports the DLO to the workspace of Pandas, where its 3D shape is estimated using the approach described in Section~\ref{subsec:handover}.
One Panda robot moves toward the DLO for grasping, prompting the UR10e to release it after the Pana co-grasped it. 
Following this, the Panda transport the DLO to the designated fixtures and secures it incooperation with the second panda within the clip.
During this process, only a few failures were observed. 
The most common one occurs when the initial grasping position of bin picking is too close to the DLO’s end, leaving insufficient space to secure it properly in the fixture.
For more information about the assembly process, please refer to the accompanying video and our project website: ~\href{https://kejiachen.github.io/DLOAssembly/}{https://kejiachen.github.io/DLOAssembly/}

\section{Conclusion}
This paper presents a comprehensive framework for the automated assembly of DLOs, including an industrial-grade supply methodology. 
The system is based on an object-centric planning approach.
Leveraging enhanced multi-modal perception, the system effectively tracks and adapts to the dynamic properties of DLOs, enabling precise coordinated manipulation. 
A key feature of the framework is a structured handover mechanism that ensures seamless transfer between robots, facilitating reliable placement along predefined trajectories. 
The proposed approach is validated through real-world robotic experiments, demonstrating its feasibility and potential for advancing automation in DLO assembly. 

Despite its effectiveness, the framework still has several limitations.
One primary limitation is its sensitivity to depth data inaccuracies. 
Noisy or imprecise depth measurements can lead to erroneous point prompts, ultimately affecting segmentation quality and grasping pose estimation.
Potential improvements include refining post-processing techniques to filter out non-target structures (e.g., connectors or bin edges) and exploring alternative depth-sensing solutions with higher accuracy.
Additionally, the spatial limitations of the experimental setup restricted certain actions.
During bin picking, motion primitives could not fully prevent contact of the picked DLO with the environment, leading to manipulation errors. 
Optimizing the workspace conditions and expanding the set of disentanglement strategies could mitigate these issues. 
Finally, the current handover mechanism limits the grasp pose to keep a certain distance to the DLO's end, so sufficient space can be reserved for mounting.
Introducing strategies to adjust the grasping point after initial grasping using collaboration between two hands could eliminate this problem.
Furthermore, the proposed pipeline does not yet encompass all stages of the industrial value chain.

In future work, we aim to extend the proposed method to integrate the connector mating process into the assembly pipeline, which connects DLOs with industrial-standard connector to other fixtures.
\bibliographystyle{IEEEtran}
\bibliography{references}

\balance

\end{document}